\pdfoutput=1
\documentclass[11pt]{article}

\usepackage{ACL2023}
\usepackage{times}
\usepackage{latexsym}
\usepackage[T1]{fontenc}
\usepackage[utf8]{inputenc}
\usepackage{microtype}
\usepackage{inconsolata}
\usepackage{mathtools}
\usepackage{amsmath}
\usepackage{amsfonts}

\title{Plausibility Vaccine: Injecting LLM Knowledge for Event Plausibility}

\author{Jacob Chmura \\
  Mila, McGill University \\\And
  Jonah Dauvet \\
  McGill University \\\And
  Sebastian Sabry \\
  McGill University \\
}

\begin{document}
{\makeatletter\acl@finalcopytrue
  \maketitle
}

\begin{abstract}

Despite advances in language modelling, distributional methods that build semantic representations from co-occurrences fail to discriminate between plausible and implausible events. In this work, we investigate how plausibility prediction can be improved by injecting latent knowledge prompted from large language models using parameter-efficient fine-tuning. We train 12 task adapters to learn various physical properties and association measures and perform adapter fusion to compose latent semantic knowledge from each task on top of pre-trained AlBERT embeddings. We automate auxiliary task data generation, which enables us to scale our approach and fine-tune our learned representations across two plausibility datasets. Our code is available at \url{https://github.com/Jacob-Chmura/plausibility-vaccine}

\end{abstract}

\section{Introduction}
Semantic plausibility studies discrimination between possible and impossible events \cite{wang-etal-2018-modeling}. From a machine learning perspective, the canonical task is a binary classification problem that decides whether subject-verb-object (s-v-o) triplets are plausible (e.g. man-ate-dinner) or implausible (e.g. bucket-drank-phone). Embedded within this learning problem are many linguistic \cite{ko-etal-2019-linguistically, porada-etal-2021-modeling, plause} challenges stemming from sparsity in language: most plausible events will not be witnessed. So, distributional methods such as transformer models tend to favour representations that infer event \textit{probability} \cite{lo-etal-2023-functional, lprob, kang-choi-2023-impact}. Several lines of work \cite{wang-etal-2018-modeling, porada-etal-2019-gorilla} investigate methods to induce knowledge such as object size, which is highly coupled with plausibility label \cite{elephant}. We aim to improve upon this body of work by increasing the fidelity of auxiliary datasets and using parameter-efficient fine-tuning \cite{pfeiffer-etal-2020-mad} to integrate latent representations without suffering from catastrophic forgetting \cite{pfeiffer-etal-2021-adapterfusion}. Existing techniques rely on human annotations. Motivated by recent literature on information extraction \cite{kwak-etal-2024-classify, wang-etal-2024-soft-knowledge} and instruction-tuned prompting \cite{blevins-etal-2023-prompting}, we posit that annotations can be automated by prompting large language models (LLMs) to bootstrap latent knowledge.

\begin{figure}[h!]
\centering
\includegraphics[width=0.35\textwidth]{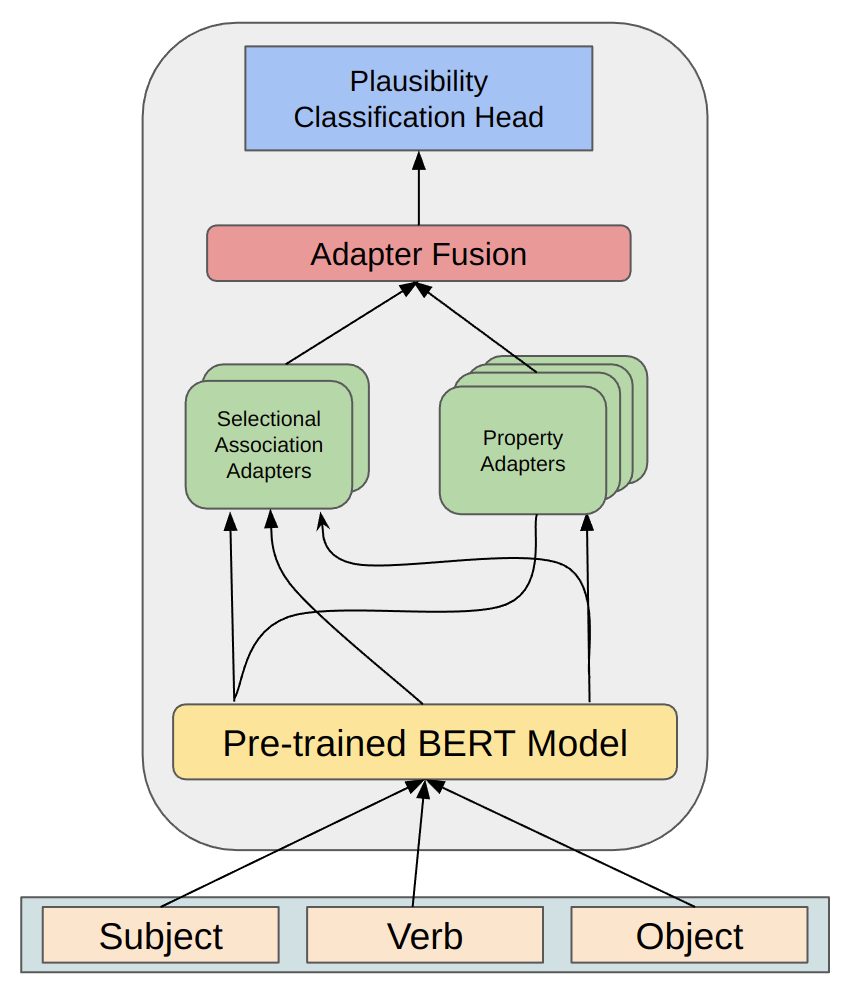}
\caption{Our high-level model architecture}
\label{fig:arch}
\end{figure}

Selection association \cite{Jurafsky:2009:SLP:1214993} reflects the tendency for words to \textit{semantically} constrain other words. We postulate that grounding verb-subject and verb-object entities with selection association signals can benefit plausibility prediction. This inspires our hypothesis:

\begin{center}    
\emph{Parameter efficient fine-tuning on LLM-generated property datasets with selection association will boost the performance of transformer-based models on plausibility classification.}
\end{center}

To test this claim, we augment the unique landmarks property datasets described in \cite{wang-etal-2018-modeling} with additional properties and selections association signals. We automate our data construction and labelling with GPT-4o \footnote{https://openai.com/index/hello-gpt-4o/} and fine-tune transformer-based language models with adapters \cite{pfeiffer-etal-2020-mad}. We train a classification layer from our learned representations on two plausibility datasets and evaluate the contribution of each task adapter to downstream performance. Overall, we find that property-based task adapters boost performance and adding selectional association adapters marginally improves performance.

\section{Related Work}

Semantic plausibility has been thoroughly investigated across various paradigms. \cite{wang-etal-2018-modeling} enhance a classifier with manually elicited entity properties including \textit{sentience}, \textit{mass-count}, \textit{phase}, \textit{size}, \textit{weight} and \textit{rigidity}. The authors utilize pre-trained GloVe vectors \cite{pennington-etal-2014-glove} to embed their s-v-o triples and train an MLP that integrates the external knowledge through a featurization scheme that directly encodes the relative physical attributes of the subject and object tokens. The differences in our work are threefold. First, based on their error analysis, we expand upon their property categories by including temperature, shape, texture, opacity and mobility.
Moreover, the authors do not inject any signal into the verb token, whereas we compute pairwise selectional association scores between verb-subject and verb-object tokens. Second, we automate our data labelling using the GPT-4o model rather than relying on human annotations. Third, we overhaul their GloVe vectors with pre-trained embeddings from the BERT model family \cite{chiang-etal-2020-pretrained}. We inject knowledge from various datasets using parameter-efficient fine-tuning adapter techniques \cite{pfeiffer-etal-2020-mad}, which enables us to transfer latent semantic understanding \textit{without relying on property labels} at test time. 

\cite{porada-etal-2019-gorilla} extended \cite{wang-etal-2018-modeling} by considering transformers. They showed that pre-trained BERT is effective at plausibility classification and outperformed previous methods. The authors explore a more fundamental problem with semantic plausibility under supervised learning: performance is highly coupled to the vocabulary coverage in the training set. They proposed a self-supervised model to combat this issue and found promising results. This work bolstered our motivation to automate dataset generation, which can be seen as a form of self-supervision.

\begin{table}
    \centering
    \begin{tabular}{||c|c||}
        \hline
        Category & Divergence \\
        \hline
        Weight  & 7.76\%  \\
        Rigidity & 7.30\% \\
        Phase & 1.83\% \\
        Sentience & 3.20\% \\
        Size & 4.57\% \\
        \hline
    \end{tabular}
    \caption{\% of samples that differ between \cite{wang-etal-2018-modeling} hand-annotated data and LLM generated labels}
    \label{tab:my_label}
\end{table}

Similar in spirit to our work, \cite{eichel-schulte-im-walde-2024-multi} presents a multi-task learning approach to semantic plausibility based on adapters. The authors categorize over 50 pre-trained adapters for various language tasks, such as named entity recognition and sentiment analysis, and then evaluate how these adapters improve plausibility prediction. Their results suggest that knowledge from different language tasks does not substantially improve and may even hurt performance. Our work differs in that we train our adapters from scratch using a combination of physical properties and semantic association rather than language tasks. Also, our data generation pipeline enables us to scale adapter pre-training. Finally, we explore additional parameters in our adapter fusion stack and test a broader range of backbone transformer models, whereas \cite{eichel-schulte-im-walde-2024-multi} only consider RoBERTa \cite{zhuang-etal-2021-robustly}.

\section{Datasets  \footnote{Our datasets are available \href{https://github.com/Jacob-Chmura/plausibility-vaccine/tree/master/data}{here.}}}

\subsection{Property Data}

\begin{figure}[h!]
\centering
\includegraphics[width=0.40\textwidth]{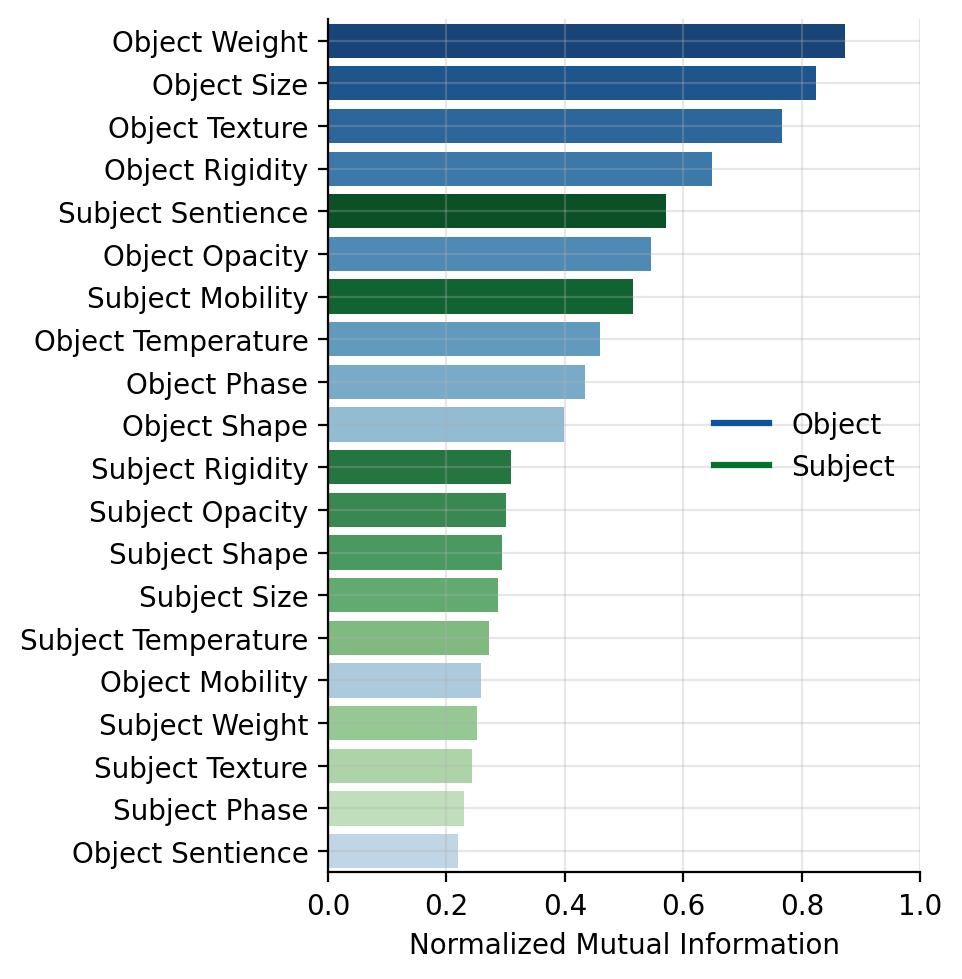}
\caption{Normalized Mutual Information Between Properties and Plausibility}
\label{fig:mi}
\end{figure}

We used GPT-4o to compile a list of 1,000 \footnote{Developing the complete list in a single prompt resulted in unreliable responses, so we generated them in batches.} distinct items that we categorize based on various properties. The first five categories—Size, Weight, Sentience, Physical Phase, and Rigidity—were drawn from \cite{wang-etal-2018-modeling}. We added Temperature, Opacity, Shape, Texture, and Mobility to enhance the dataset and improve embedding representation.  Each item was assigned a numerical value corresponding to its category based on predefined reference points. Details on our numeric assignments can be found in \href{https://github.com/Jacob-Chmura/plausibility-vaccine}{our repository}. The labelling was also automated using GPT-4o with prompts like \textit{"Categorize each item according to its size with respect to the reference points: ant, cat, person, jeep, stadium."}. Figure~\ref{fig:mi} highlights that object features like weight, size, and texture exhibit higher mutual information with plausibility values. Interestingly, subject properties have less utility than their object-specific counterparts.

\subsection{Selectional Association Data}
\begin{figure}[hbt!]
\centering
\includegraphics[width=0.7\linewidth]{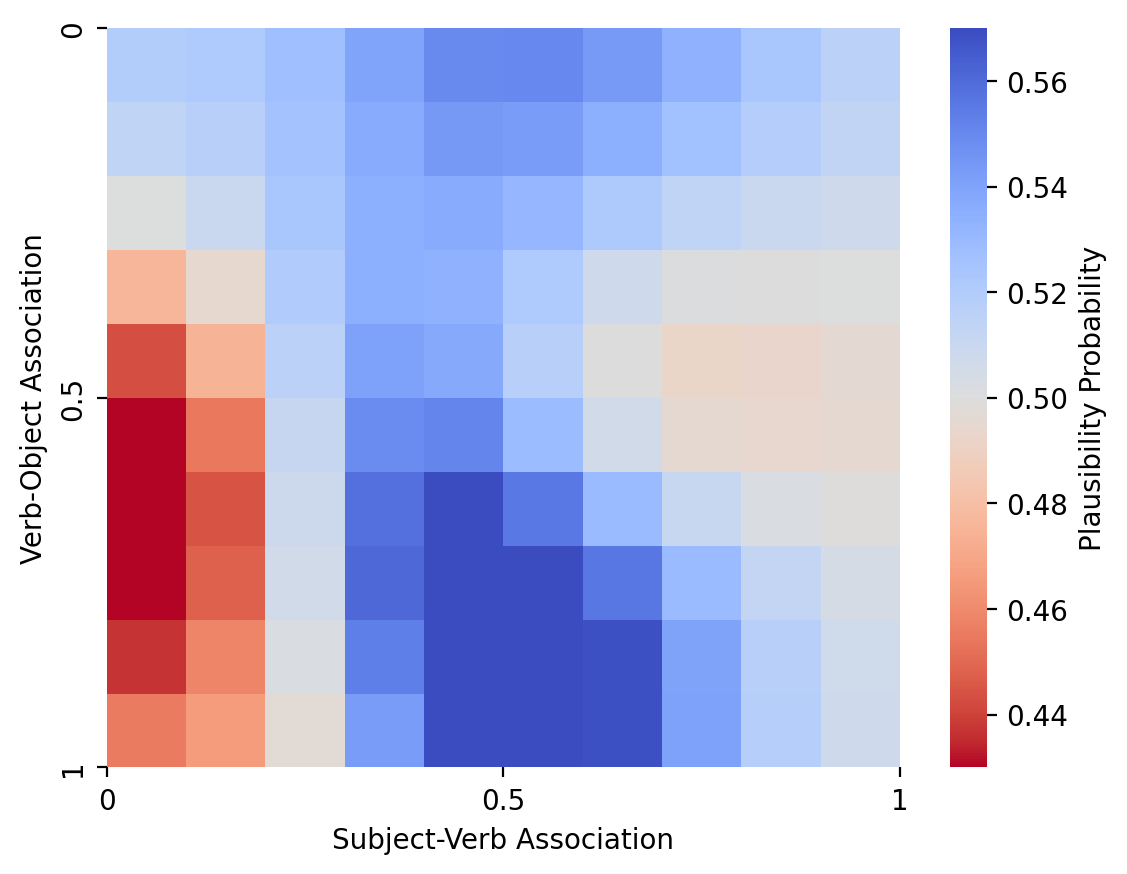}
  \caption{Joint correlation between SV,OV association and plausibility on \textit{PEP-3K} validation dataset.}
  \label{fig:joint_corr}
\end{figure}

Previous work did not encode verb usage in s-v-o triples \cite{wang-etal-2018-modeling}. Suppose we would like to classify the plausibility of \textit{monkey-pluck-apple}. We would require some property of a verb that restricts selection. The verb \textit{pluck} semantically requires a tangible object. How can we enforce or estimate this restriction? This question has been explored under \textit{selection association} \cite{Jurafsky:2009:SLP:1214993}. First, the selection preference of a verb is defined as the information the verb expresses about a possible semantic class:
\begin{equation}
    S_{R}(v) = D(P(c|v) || P(c))
\end{equation}
where $D(P || Q)$ is the Kullback-Leibler divergence. Selection association is a probabilistic measure of association for a verb and class:
\begin{equation}
    A_R(v,c) = \frac{1}{S_{R}(v)}P(c|v)\log{\frac{P(c|v)}{P(c)}}
\end{equation}
Thus, given a corpus, we can measure how much the verb constrains the subject or object \cite{brockmann-lapata-2003-evaluating}. Figure~\ref{fig:joint_corr} describes the joint plausibility probability relative to object and subject-verb associations. When associations are low, plausibility is statistically lower. This shows that selection association can be a determiner of plausibility and that object-verb and subject-verb association should be used jointly. Further investigations into the association and plausibility correlation describe non-linear relationships warranting adapters to capture this relationship. \footnote{Relevant figures are available at  \href{https://github.com/Jacob-Chmura/plausibility-vaccine}{our repository}.} We compute association scores across all verb-subject and verb-object pairs on the \textbf{20Q} and \textbf{PEP-3k} datasets.
\vspace{-10pt}  % MANUAL: Reduce space after the table

\subsection{Plausibility Data}

\textbf{PEP-3K} \cite{wang-etal-2018-modeling} A crowdsourced \textbf{P}hysical \textbf{E}vent \textbf{P}lausibility dataset which consists of 3062 s-v-o triples built from a vocabulary of 150 verbs and 450 nouns, manually rated for physical plausibility. We follow the validation and test splits of \cite{porada-etal-2019-gorilla}.

\textbf{20Q} \footnote{https://github.com/allenai/twentyquestions} A crowdsourced question-answer judgment dataset adapted for binary plausibility classifications by \cite{porada-etal-2021-modeling}. The data consists of 5096 s-v-o triples. We follow the validation and test splits of \cite{porada-etal-2021-modeling}.

Both plausibility datasets additionally include word senses. Table~\ref{tab:wsd} shows the distribution of sense rank across both datasets according to Wordnet \cite{miller-1994-wordnet}. Since most tokens belong to the most frequent rank, we posit that distributional embeddings implicitly capture the correct sense in most cases \cite{wsd}.
\begin{table}
\begin{tabular}{lrrrr}
\hline
Dataset          &   Rank 1 &   Rank 2 &   Rank 3 &   Rank $\geq$ 4 \\
\hline
 PEP         &           0.62 &           0.16 &           0.09 &    0.14 \\
 20q &           0.77 &           0.12 &           0.06 &    0.06 \\
\hline
\end{tabular}
\caption{Word Sense Rank Distribution according to Wordnet v3.0 \url{https://wordnet.princeton.edu/}}
\label{tab:wsd}
\end{table}

\section{Method}
Our high-level architecture is given in Figure~\ref{fig:arch}.

\subsection{Transformer Backbone}

We utilize the \texttt{AdapterHub} framework \cite{pfeiffer-etal-2020-adapterhub}, which hosts state-of-the-art language models and enables fine-tuning on top of the HuggingFace transformers library \footnote{https://github.com/huggingface/transformers}. Specifically, we use the \texttt{Albert-base-v2} model checkpoint \cite{chiang-etal-2020-pretrained} and keep the parameters frozen. The s-v-o triplet is tokenized using \textit{SentiencePiece} \cite{kudo-richardson-2018-sentencepiece}.

\subsection{Adapter Pretraining and Fusion}

We leverage the \texttt{Adapters} library \cite{poth-etal-2023-adapters} to pre-train a task adapter for each physical property (size, weight, etc.) and both subject-verb and verb-object associations. We use sequential bottleneck adapters \cite{pfeiffer-etal-2020-mad} at every layer of the transformer backbone with \texttt{ReLu} nonlinearities and reduction factors of 16. All other adapter parameters default to the \texttt{Adapters} library. 

Adapters are trained using \textit{Adam} \cite{2015-kingma} with a learning rate of $1e^{-4}$ on datasets of the form $\mathcal{D}_{prop} = \{(\textbf{x}_i, y_i)\}_{i = 1, ..., N}$
where $\textbf{x}_i \in \mathbb{R}^d$ is the d-dimensional s-v-o feature representation from the transformer. For property based adapters, the target $y_i \in \{1, ..., 5\}$ is the one-hot property encoding. The target $y_i \in [0, 1]$ is the association score between the verb-subject or verb-object tokens for association adapters. The optimization objective is to find parameters $\theta$ that minimize the cross entropy loss $\mathcal{L}_{ce}$ (resp. mean squared error $\mathcal{L}_{mse}$ on the property (resp. association) datasets. After pre-training all 12 adapters, we employ adapter fusion \cite{pfeiffer-etal-2021-adapterfusion} to compose the learned representations.

\subsection{Plausibility Finetuning}

We freeze the learned embeddings and stack a single-layer MLP classifier on our representation. The training procedure uses the same optimization parameters as adapter pre-training (5 epochs, \textit{Adam} optimizer, learning rate $1e^{-4}$) and minimizes the binary cross entropy $\mathcal{L}_{BCE}$ on the plausibility data.

\subsection{Experiment Setup}

We partition the held-out test dataset into five disjoint groups and report mean and standard deviation performance across these partitions. In our experiments, \texttt{N} corresponds to a baseline model which fine-tunes a plausibility MLP classification layer directly on top of our transformer. The \texttt{P} setup integrates property-based adapters, and \texttt{P + V} integrates property-based and verb-based selection association adapters. Methods labelled with \textit{Combined} were trained on 20q and PEP datasets. Otherwise, each experiment was trained on the training split of the evaluated dataset.
\section{Results}

Due to limited space, we could not discuss all our experiments and ablations. We implore readers to visit \href{https://github.com/Jacob-Chmura/plausibility-vaccine}{our repository} for further details and results.

\subsection{Adapter Pre-Training Performance}

We report accuracy and macro-F1 for property adapters and mean-squared error on their held-out dataset in table ~\ref{tab:adaptperf}. As expected, performance varies with the task. Some properties like shape and rigidity are relatively ambiguous in their categorization, which we hypothesize is responsible for their subpar performance. Still, we observe strong results for multi-class classification for various properties. By design, our adapter fusion methodology is non-sensitive to adapter bias: mispredictions do not propagate to downstream error since we solely use the \textit{representations} from the adapter-pre-training phase rather than the prediction heads themselves.

\begingroup
\setlength{\tabcolsep}{4pt} 
\begin{table}[hbt!]
\centering
\begin{tabular}{llll}
\hline
 Property Task        & Accuracy     & F1      \\
\hline
Mobility    & 0.83 ± 0.12  & 0.6 ± 0.12\\
Phase       & 0.82 ± 0.027 & 0.33 ± 0.096\\
Sentience   & 0.69 ± 0.062 & 0.23 ± 0.039\\
Opacity     & 0.53 ± 0.097 & 0.27 ± 0.056\\
Size        & 0.51 ± 0.053 & 0.21 ± 0.016\\
Texture     & 0.4 ± 0.078  & 0.16 ± 0.058\\
Weight      & 0.36 ± 0.092 & 0.18 ± 0.059\\
Temperature & 0.34 ± 0.039 & 0.21 ± 0.041\\
Rigidity    & 0.31 ± 0.082 & 0.18 ± 0.057\\
Shape       & 0.30 ± 0.046  & 0.16 ± 0.022\\
\hline
 Association Task               & MSE           \\
\hline
 verb\_subject\_score & 0.07 ± 0.01 \\
 verb\_object\_score  & 0.05 ± 0.01 \\
\hline
\end{tabular}
\caption{Adapters Pre-training Performance}
\label{tab:adaptperf}
\end{table}
\endgroup

\subsection{Downstream Plausibility Performance}

Our main results are summarized in table \ref{tab:perf}, which shows accuracy across various configurations. Adding property-based adapters (\texttt{P}) improves upon direct plausibility fine-tuning (\texttt{N}) by 8\% and 2\% in 20q and PEP datasets, respectively. Fusing selectional association adapters further enhances performance by 2-3\%. Training on both datasets (\texttt{Combined}) does not substantially help. We hypothesize that this is explained by the relatively small amount of plausibility data compared to pre-training data.

\begin{table}
\begin{tabular}{rll}
\hline
    Task   & Adapter Fusion           & Accuracy    \\
\hline
  20q    & N & 0.55 ± 0.028   \\
  20q    & P & 0.63 ± 0.019   \\
  20q    & P + V & \textbf{0.65 ± 0.027} \\
  20q    & N (Combined)              & 0.56 ± 0.016   \\
  20q    & P (Combined) & 0.64 ± 0.02    \\
  20q    & P + V (Combined) & \textbf{0.65 ± 0.024}  \\
\hline
  PEP    & N & 0.5 ± 0.018  \\
  PEP    & P & 0.52 ± 0.004  \\
  PEP    & P + V & \textbf{0.54 ± 0.015}  \\
  PEP    & N (Combined) & \textbf{0.54 ± 0.043}  \\
  PEP    & P (Combined) & \textbf{0.54 ± 0.022}  \\
  PEP    & P + V (Combined) & \textbf{0.54 ± 0.037}  \\
\hline
\end{tabular}
\caption{Plausibility Performance}
\label{tab:perf}
\end{table}

\subsection{Ablations}

We test whether performance is improved when using larger transformers (BERT, RoBERTa). Interestingly, table ~\ref{tab:transformer_abl} shows that RoBERTa backbone performs degrades relative to AlBERT on the 20q dataset. We also tested different adapters and found that our approach is robust to bottleneck size.

\begin{table}
\centering
\begin{tabular}{rll}
\hline
    Ablation  & Task   & Accuracy \\
\hline       
  AlBERT & 20q  &  0.65 ± 0.024  \\
 BERT & 20q    &  \textbf{0.66 ± 0.018} \\
 RoBERTa & 20q    & 0.63 ± 0.02 \\
 \hline
 AlBERT & PEP    & 0.54 ± 0.037 \\
 BERT & PEP    & \textbf{0.57 ± 0.02} \\
 RoBERTa & PEP    & 0.56 ± 0.023 \\
\hline
\end{tabular}
\caption{Transformer Backbone Ablation Showing Plausibility Performance}
\label{tab:transformer_abl}
\end{table}

\section{Discussion and Limitations}
We conclude that there is a net improvement from adding new physical categories and using parameter-efficient fine-tuning techniques to inject specific property knowledge into pre-trained embeddings. While we demonstrated that our LLM-labeled corpus has low divergence from Wang et al.'s dataset, only 30\% of our items overlapped with theirs, leaving 70\% of the labels unverified. Additionally, the lack of an API key for GPT-4o posed challenges for scaling our approach. However, future work could address these challenges with sufficient financial resources.
\section{Conclusion and Future Work}

This paper introduced a method to improve semantic plausibility using adapter fusion with knowledge injected from GPT-4o. We improved upon directly fine-tuned BERT embeddings by combining property-based adapters with selectional association. We highlight several directions for future work. We implore replication of our findings with other language models to investigate how dataset quality impacts downstream performance. Also, researchers should explore techniques such as prompt chaining \cite{promptchain} to improve the quality of proxy datasets or RAG \cite{graphrag} applications to ground these datasets in trusted knowledge bases. An interesting research direction would be to find a minimal set of adapters that improve plausibility without perturbing performance in the multi-task setting. Injecting semantic understanding in this way could enhance reasoning capabilities in language models that are prone to hallucinate \cite{hallucinate}.

\section{Statement of Contributions}
\subsection{Jacob Chmura}

I took ownership of our codebase, wrote the training and evaluation procedure, implemented baselines, adapter fine-tuning, adapter fusion, and model checkpointing, and retrieved the plausibility data. I implemented cross-validation and scripts that run the system end-to-end. I also created scripts to generate the charts in this report and added ablation analytics. I wrote the abstract and sections 2, 3.3, 4, 5, 7.

\subsection{Jonah Dauvet}

I designed, implemented and validated property data labelling and wrote code to assess the overlap between our datasets and \cite{wang-etal-2018-modeling}. I authored sections 3.1, 6.

\subsection{Sebastian Sabry}
I devised and implemented selection association, created our dataset for verb association, and researched semantic plausibility papers. I authored sections 2, 3.2, and the introduction.

\clearpage
\bibliography{0-main}
\bibliographystyle{acl_natbib}

\end{document}